%% file: main.tex
\pdfoutput=1
\documentclass{article} % NeurIPS style expects 10pt article; omit [12pt]

\usepackage{neurips}   % <- your neurips.sty
\nipsfinalcopy         % comment this out if you want anonymous authors + gray ruler

\usepackage[utf8]{inputenc}
\usepackage[T1]{fontenc}
\usepackage{times}
\usepackage{microtype}

\usepackage{graphicx}
\usepackage{import}
\usepackage{algorithm}
\usepackage{algpseudocode}

\usepackage[hidelinks]{hyperref}
\usepackage{xurl}

\usepackage{amsmath}
\usepackage{amssymb}
\usepackage{array}
\usepackage{booktabs}
\usepackage{float}

\usepackage[
  backend=biber,
  style=numeric,
  sorting=nyt,
  maxbibnames=10,
  maxcitenames=2
]{biblatex}

\AtBeginDocument{\let\cite\parencite}

\title{When Context Sticks: Studying Interference in In-Context Learning}

\author{\normalfont
  Hanna Rød\textsuperscript{*}
  \and Dagny Streit\textsuperscript{*}
  \and Nils Valseth Selte\textsuperscript{*}
  \and Justin Li\textsuperscript{*}
}

\begin{document}

\maketitle
\begingroup
\renewcommand{\thefootnote}{*}
\footnotetext{Equal contributors. Code available at: \url{https://github.com/nilsvselte/icl-context-stickiness}}
\endgroup
\begin{abstract}
This paper investigates context stickiness in in-context learning (ICL), a phenomenon where earlier examples in a prompt interfere with a transformer’s ability to adapt to later tasks. Using synthetic regression tasks over linear and quadratic functions, we examine how models trained under sequential, mixed, and random curricula handle abrupt task switches during inference. By sweeping over structured combinations of misleading linear examples followed by recovery quadratic examples, we quantify how prior context biases prediction error and how quickly models realign. Our results show strong evidence of persistent interference: more preceding linear examples reliably degrade quadratic predictions, while additional quadratic examples reduce error but with diminishing returns. We further find that training curricula significantly modulate resilience, with sequential training on the target function class yielding the fastest recovery, and surprisingly, random training producing the least robust behavior.

\end{abstract}

\import{./Sections/}{Introduction}
\import{./Sections/}{Litterature_review}
\import{./Sections/}{Key_questions}
\import{./Sections/}{Hypothesis}
\import{./Sections/}{Methods}

\import{./Sections/}{Hyperparam_sweep}

\import{./Sections/}{Results}

\import{./Sections/}{Conclusion}

\newpage
\printbibliography

\appendix

\section{Acknowledgments}
We thank the UC Berkeley staff and fellow students for feedback that improved the paper. In particular, we thank Prof. Anant Sahai for providing feedback and arXiv endorsement. We also want to thank Thomas Hasvold Adam and Marcel Rød for their assistance, as well as Michael Equi for helping us brainstorm the idea behind the paper.

\section{AI Disclosure}
% \label{app:ai-disclosure}
We used generative AI to help with the figure plotting and modifying some of the inference code. We also used it to help edit our report.
    
\end{document}

%% file: Sections/Introduction.tex
\section{Introduction}
In-context learning (ICL) is a model's ability to use examples in the prompt to make accurate predictions without updating its weights \cite{brown2020language}.

As context windows grow, it becomes increasingly important to understand how transformers use and misuse information provided in context during inference \cite{liu2023lostmiddlelanguagemodels, bertsch2025incontextlearninglongcontextmodels}. Prior work has also shown that ICL is sensitive to prompt order, formatting, and contextual structure \cite{lu2022fantasticallyorderedpromptsthem, min2022rethinkingroledemonstrationsmakes, zhao2021calibrateuseimprovingfewshot}. Recent work further suggests that in long-context regimes, the challenge may shift from carefully selecting a small set of demonstrations to effectively utilizing and filling the available context with sufficient examples \cite{baek2025revisitingincontextlearninglong}.

This distinction matters because strong in-context performance does not necessarily imply that a model is fully learning the new task from the examples alone. Prior work has shown that part of the gain may come from recognizing the task structure or prompt schema itself, rather than fully inferring the input-output rule from context \cite{pan2023incontextlearninglearnsincontext}. Our setting therefore studies a harder question: when the contextual evidence changes abruptly, how quickly can the model abandon an earlier inferred task and adapt to a new one?

By sweeping over structured combinations of misleading ICL examples followed by recovery examples, we isolate the cost of task switching at the level of the learned inner-loop adaptation procedure. More concretely, we study how much misleading prior context biases prediction error and how many post-switch examples are needed for the model to recover.

We find that even a small amount of interference can significantly degrade ICL performance in task-specific trained transformers. We further find that training curricula affect the relationship between misleading context and recovery. Lastly, we provide a code repository to support further experimentation on these findings.

We study this question in a controlled synthetic setting, using small self-trained transformers on function-class regression tasks, so our results should be interpreted as mechanistic evidence about interference under task switching rather than as a direct quantitative claim about frontier pretrained LLMs.

%% file: Sections/Litterature_review.tex
\section{Literature Review}
\subsection{Definition and Introduction}
We define ICL following Brown et al. and later survey work by Dong et al. \cite{brown2020language, dong2024surveyincontextlearning} as:
\textit{In-context learning is a paradigm that allows language models to learn tasks given only a few examples in the form of demonstration.}

Transformers exhibit strong ICL abilities. Scaling up the parameter counts of large language models has led to increasingly task-agnostic few-shot learning capabilities \cite{kaplan2020scalinglawsneurallanguage,brown2020language}. With GPT-3, it was shown that, in some cases, a model using only in-context examples can sometimes approach or reach competitiveness with prior fine-tuned approaches \cite{brown2020language}. This work also introduced the modern view of ICL as an implicit meta-learning process composed of two distinct loops. The outer loop consists of standard gradient-based training, where the model's parameters are optimized over a broad distribution of tasks. In contrast, the inner loop occurs entirely during inference. In the inner loop the model utilizes the provided examples to adapt its internal representations to a specific task without updates to its weights or biases \cite{oswald2022mesa}.

Later in this literature review, we will also see how these ideas have influenced the broader goal of developing continuously learning models.

\subsection{Prompt Sensitivity}
These ICL behaviors can also be sensitive to prompt construction. Example order can have a significant effect on performance \cite{lu2022fantasticallyorderedpromptsthem, zhao2021calibrateuseimprovingfewshot}, and in some cases, ordering can matter more than the specific choice of examples \cite{li2025ordermattersrethinkingprompt}. On classification and multiple-choice tasks, replacing ground-truth labels with random labels often causes only a modest performance drop. This suggests that demonstrations may help by identifying the label space and prompt format rather than only by providing correct supervision \cite{min2022rethinkingroledemonstrationsmakes}.

\subsection{Templates for Small-Scale ICL Experiments}
In our experiments, we adopt the template-based synthetic setups introduced by Garg et al. In this framework, transformers are trained on sequences of input–output pairs of the form $(x, f(x))$, where the pairs are drawn from different function classes. Garg et al. demonstrate that, under this setup, small transformers can be trained to in-context learn linear functions, sparse linear functions, small neural networks, and decision trees, achieving performance that is comparable to that of task-specific estimators designed for these problems \cite{garg2022template}. Throughout our work, we employ the same template when constructing and running our experiments.

\subsection{Algorithmic and Mechanistic Views of ICL}

Later work proposes an algorithmic account of ICL, showing that transformers can in some settings recover behaviors resembling standard learning algorithms and viewing the forward pass as a learned optimization procedure \cite{akyürek2023learningalgorithmincontextlearning}.

Von Oswald et al.\ demonstrate that a single self-attention layer can be explicitly constructed to perform a step of gradient descent (GD) on the squared loss. They further show that transformers trained via standard optimization can rediscover the artificially constructed weights \cite{oswald2022mesa}.

Dai et al. extend this optimization-based perspective to pretrained GPT-style language models, arguing that ICL can be understood as implicit finetuning, where the model acts as a meta-optimizer that produces update-like behavior through forward computation \cite{dai2023gptlearnincontextlanguage}.

Bai et al.\ extend this view by proving that suitably parameterized transformers can approximate several steps of gradient descent on a broad class of smooth convex objectives, and can in-context implement standard estimators such as least squares and regularized regression methods. They further show that a single model can perform in-context algorithm selection, effectively choosing between different base procedures based on statistics of the prompt \cite{bai2023iclalgorithms}. Building on these findings, we treat the transformer as a learned inner-loop optimizer. By viewing the model as an algorithm that actively updates its hypothesis with every new token, we can quantify how resilient this optimization process is when the 'training data' in the context suddenly changes distribution.

Xie et al. provide a complementary perspective, explaining ICL as a form of implicit Bayesian inference rather than purely as learned gradient-based optimization \cite{xie2022explanationincontextlearningimplicit}.

On the mechanistic side, Olsson et al. identify induction heads as a circuit-level mechanism that can support context-dependent pattern matching and retrieval \cite{olsson2022incontextlearninginductionheads}. Todd et al. propose a complementary account, showing that a small number of attention heads can transport compact task representations, which they call function vectors, across a range of in-context learning tasks \cite{todd2024functionvectorslargelanguage}. More recently, Yin and Steinhardt compare induction heads and function-vector heads across multiple language models, finding that few-shot in-context learning in larger models depends primarily on function-vector heads, while also suggesting that induction heads may serve as a precursor during training \cite{yin2025attentionheadsmatterincontext}.

\subsection{Conflicting Context and Task Switching}

However, despite these strong ICL capabilities, models are still vulnerable to interference and task switching. Coleman et al.\ study in-context interference in chat-based LLMs using a benchmark derived from the bAbI dataset. They construct sequences of short stories and questions such that a Vicuna-13B model can answer all questions perfectly when each story is presented in its own context, but then incrementally add more stories into a single growing context. As the number of stories increases, accuracy drops from 100\% with one story to around 75\% with eight, even though no model weights are updated. Adding more context or information can thus cause the model to forget or overwrite previously acquired knowledge \cite{coleman2023interference}.

Coleman et al. relate this phenomenon to catastrophic forgetting, but emphasize that the interference here occurs at the level of the context window rather than parameter updates. They also test simple mitigation strategies, finding that summarizing earlier stories does not prevent degradation, whereas removing older stories and keeping only a fixed-size buffer can stabilize performance. This indicates that how information is organized and pruned in the prompt can strongly affect in-context behavior \cite{coleman2023interference}. In our work, we build on this perspective by using controlled synthetic tasks to study how conflicting examples and explicit task switches interfere with a transformer's learned inner-loop algorithm. We will refer to this phenomenon as context stickiness.

\subsection{Task Mixtures and Sampling Strategies}
The task-switching and algorithmic views both lead to the question of how different outer-loop learning strategies affect inner-loop learning. If the outer loop is responsible for learning the inner meta-learning optimizer, then the curriculum presented during training should directly dictate the capabilities the model acquires to solve tasks. Relatedly, Raventós et al. show that pretraining task diversity can change the kind of in-context learner that emerges, suggesting that robustness under task mixtures may also depend on the diversity of the training task distribution \cite{raventós2023pretrainingtaskdiversityemergence}.

While most early function-class studies trained on a single task family, Bhasin et al. investigate how different training curricula over multiple tasks affect ICL \cite{bhasin2024samples}.

Using the template of Garg et al., Bhasin et al. train transformers on a mixture of different tasks with three sampling strategies: purely sequential, uniform random, and mixed. The mixed curriculum gradually introduces harder tasks while continuing to sample earlier ones. Their results show that these choices have a substantial impact on ICL performance, with the mixed curriculum tending to perform better and converging on tasks where the other curricula fail \cite{bhasin2024samples}. In fact, it has been shown that training on multiple ICL tasks simultaneously shortens loss plateaus \cite{kim2025taskdiversity}.

Our paper shows that these training curricula also affect how robust models are to conflicting context.

\subsection{A Modern View on Learning Systems}

Recent work introduces Nested Learning as a general framework for interpreting deep learning models as collections of interacting inner- and outer-loop learners operating at different timescales \cite{behrouz2025nestedlearning}. In this perspective, the phenomenon of catastrophic interference also describes how quickly an inner-loop learner overwrites information in its context window.

While our experiments utilize a standard transformer rather than the specialized multi-timescale architectures proposed by Behrouz et al., they probe a critical slice of this broader picture: the fragility of a single inner loop. By quantifying how conflicting examples and task mixtures perturb the model's predictions, we provide empirical evidence of the limits of standard inner-loop optimization. Our results effectively map the boundary where a single-timescale learner fails.

%% file: Sections/Key_questions.tex
\section{Key Questions}
We aim to study how prior contextual exposure influences a transformer’s ability to adapt to different tasks during inference.

\begingroup
\setlength{\parskip}{0.4\baselineskip}
\setlength{\parindent}{0pt}

\textbf{Persistence of prior context}: Within an in-context sequence, does exposure to linear examples make the model's subsequent quadratic predictions less accurate compared to predictions made after seeing only quadratic examples? 

\textbf{Adaptation rate}: Within an in-context sequence, how many quadratic examples are required for the model to recover from prior linear exposure and reach a similar MSE to the quadratic-only evaluation curve?

\textbf{Influence of prior strength}: Within an in-context sequence, does increasing the number of linear prior examples affect how many quadratic examples are needed for recovery?

\textbf{Curriculum dependence}: How do persistence, adaptation rate, and influence of priors differ across models with sequential, mixed, and random curricula?

\endgroup

%% file: Sections/Hypothesis.tex
\section{Hypothesis}

\begingroup
\setlength{\parskip}{0.4\baselineskip}
\setlength{\parindent}{0pt}

\textbf{Persistence of prior context}: Within an in-context sequence, a transformer that is first exposed to linear examples will show higher MSE on subsequent quadratic predictions than a model given only quadratic examples, indicating in-context stickiness from the earlier linear context.

\textbf{Adaptation rate}: Within an in-context sequence, as more quadratic examples appear after the switch from linear to quadratic, the model's MSE on quadratic predictions should steadily decrease, eventually approaching the quadratic-only baseline.

\textbf{Influence of prior strength}: Within an in-context sequence, increasing the number of prior linear examples will increase the number of quadratic examples required to recover.

% \textbf{Context reactivation}: Within an in-context sequence, after switching from class A to class B, the model will partially retain the ability to recall and correctly apply the earlier class A.

\textbf{Curriculum dependence}: Models trained under a sequential curriculum with the target function class second should show the strongest recovery. Mixed-curriculum models should fall between these extremes, while randomly trained models should show the weakest recovery and highest error under dual inference.

\endgroup

%% file: Sections/Methods.tex
\section{Methods} \label{sec:methods}

\subsection{Data Generation}
We follow the experimental setup of ``What Can Transformers Learn In-Context? A Case Study of Simple Function Classes" \cite{garg2022template}, training a small GPT-style transformer on two function classes: two-dimensional linear ($\mathcal{F}_1$) and quadratic ($\mathcal{F}_2$) functions. Each training or inference sequence is generated by first sampling a fresh function (i.e., a weight vector) and then evaluating that function on independently drawn Gaussian inputs.

\textbf{Inputs:} For every trial we draw input vectors
\[
x \in \mathbb{R}^{2}, \qquad x \sim \mathcal{N}(0, I_2),
\]

\textbf{Linear Regression:} For each linear sequence we sample a weight vector
\[
w \in \mathbb{R}^{2}, \qquad w \sim \mathcal{N}(0, I_2),
\]
and compute outputs using
\[
y = \ x^\top w.
\]

\textbf{Quadratic Regression:}
Quadratic tasks use the same weight sampling \(w \sim \mathcal{N}(0, I_2)\), but operate on
elementwise-squared inputs. For each \(x\), let \(x^{\odot 2}\) denote the vector obtained by
squaring each coordinate. Outputs are computed as
\[
y 
= \\ \frac{(x^{\odot 2})^\top w}{\sqrt{3}}.
\]
where the \(1/\sqrt{3}\) factor normalizes the output variance, ensuring linear and quadratic tasks lie on comparable scales.

\subsection{Training Curricula}
To study how task ordering in training affects in-context learning, we adopt the three curricula introduced in ``How Does Multi-Task Training Affect Transformer In-Context Capabilities? Investigations with Function Classes" \cite{bhasin2024samples}: sequential, mixed, and random. We train the model using mini-batches. Training proceeds for $T$ steps. At each step, the curriculum selects a function class, and we generate a batch of independently sampled functions from that class.

\textbf{Sequential Curriculum:} We divide the training into two equal partitions of length $T/2$. The model trains exclusively on $\mathcal{F}_1$ during the first half of training and exclusively on $\mathcal{F}_2$ during the second half of training.
\[
f_t \sim
\begin{aligned}
  &\begin{cases}
      \mathcal{F}_1, & 1 \le t < \tfrac{T}{2}, \\[4pt]
      \mathcal{F}_2, & \tfrac{T}{2} \le t \le T.
    \end{cases}
\end{aligned}
\]

\textbf{Mixed Curriculum:} We divide the training into two halves. During the first half, tasks are drawn only from $\mathcal{F}_1$. During the second half, the model trains on a uniform mixture of the two classes. Let $\xi \sim \mathrm{Unif}\{1,2\}$.
\[
  f_t \sim
  \begin{cases}
    \mathcal{F}_1, & 1 \le t < \tfrac{T}{2}, \\[6pt]
    \displaystyle \sum_{s=1}^{2} \mathbf{1}\{\xi = s\}\,\mathcal{F}_s,
      & \tfrac{T}{2} \le t \le T.
  \end{cases}
\]

\textbf{Random Curriculum:} The model receives no structured ordering of training. At each step $t$, the function class is chosen uniformly at random from the two classes. Let $\zeta \sim \mathrm{Unif}\{1,2\}$ i.i.d.\ across $t$.
\[
  f_t \sim \sum_{s=1}^{2} \mathbf{1}\{\zeta = s\}\,\mathcal{F}_s,
  \qquad 1 \le t \le T.
\]

We exclude within-sequence task switches during training so that recovery or interference at inference reflects generalization of the learned inner-loop optimizer, rather than explicit switch training.

\subsection{Inference and Evaluation}

\noindent To evaluate how well in-context representations persist and adapt, we design an inference protocol based on a sudden shift in the underlying task distribution. We assess adaptation performance in two symmetric directions:

\begin{enumerate}
    \item \textbf{Linear $\to$ Quadratic:} The sequence begins with linear examples and switches to quadratic examples. The model must adapt to predict a quadratic query.
    \item \textbf{Quadratic $\to$ Linear:} The sequence begins with quadratic examples and switches to linear examples. The model must adapt to predict a linear query.
\end{enumerate}

\noindent For both scenarios, we sweep over context counts $n_l,n_q \in \{0,1,2,3,4,6,8,10,12,14,16,18,20\}$, restricting the total sequence length $(n_l+n_q)$ to be at most 41, matching the training setup in~\cite{garg2022template}. Note that configurations where the pre-switch count is 0 represent a standard baseline without distribution shift.

For each $(n_l, n_q)$ configuration, we run 1,000  independent batch-level trials. In every trial, we resample fresh functions for both classes (e.g., a fresh linear and a fresh quadratic function). These are used to generate the context examples, and the model is prompted to predict a held-out query drawn only from the post-switch function. We report the averaged MSE on the prediction after the context switch, summarizing the model's ability to adapt to the new function class. Algorithm \ref{alg:context-switch-eval} details this high-level inference procedure.

\begin{algorithm}
\caption{Evaluation under context switches}
\label{alg:context-switch-eval}
\begin{algorithmic}[1]

\State Define models $\mathcal{M}$ with associated checkpoints
\State Define directions $\mathcal{D} = \{\text{Linear}\!\rightarrow\!\text{Quadratic},\,
\text{Quadratic}\!\rightarrow\!\text{Linear}\}$
\State Define grid $\mathcal{G}$ of context lengths $(n_{\mathrm{pre}}, n_{\mathrm{post}})$
\State Define number of batch-level trials $R$ and batch size $B$

\For{each model $m \in \mathcal{M}$}
    \For{each direction $d \in \mathcal{D}$}
        \For{each $(n_{\mathrm{pre}}, n_{\mathrm{post}}) \in \mathcal{G}$}
            \State Initialize error list $E \gets [\,]$
            \For{$r = 1$ to $R$}
                \State Sample a batch of fresh linear functions and quadratic functions
                \State Sample context and query inputs independently from $\mathcal{N}(0, I_2)$
                \State Construct contexts $C$ with $n_{\mathrm{pre}}$ examples from the first task in $d$
                \Statex \hspace{\algorithmicindent} followed by $n_{\mathrm{post}}$ examples from the second task in $d$
                \State Sample held-out queries from the post-switch task and compute targets
                \State Run model $m$ on the contexts and queries to obtain predictions
                \State Append the batch-mean squared error to $E$
            \EndFor
            \State Compute $\mathrm{mean}(E)$ and $\mathrm{SEM}(E)$
            \State Save results for $(m, d, n_{\mathrm{pre}}, n_{\mathrm{post}})$
        \EndFor
    \EndFor
\EndFor

\end{algorithmic}
\end{algorithm}

When visualizing the results of this evaluation, each point in our plots corresponds to the averaged MSE defined above for a fixed $(n_l, n_q)$ configuration and a given training curriculum. For each configuration, we also compute the sample standard deviation of the MSE values across trials. We use the standard error of the mean (SEM) as our vertical error bars, computed as 
\[
\mathrm{SEM} = \frac{\mathrm{SD}}{\sqrt{R}},
\]
where $\mathrm{SD}$ is the sample standard deviation of the trial MSEs and $R$ is the number of trials. This choice emphasizes the uncertainty in our estimate of the expected MSE for each adaptation setting, rather than the variability of individual trials, and allows us to compare how reliably the different curricula support in-context adaptation across the range of $(n_l, n_q)$ configurations.

\noindent Training and inference required modest compute of approximately 15 hours in total on a single NVIDIA RTX 4090.

%% file: Sections/Hyperparam_sweep.tex
\section{Hyperparameter and Architecture Choice}

Our architecture and hyperparameter choices build directly on the setup of Garg et al.~\cite{garg2022template}, who use a GPT-2-style decoder-only transformer with modified input and output layers. Specifically, instead of token embeddings and tied output projection, their model uses two independent linear layers: one that maps each real-valued example $(x, y)$ into the embedding space, and one that maps the final hidden state back to a scalar prediction. We retained this structure for consistency with prior work and because it provides a clean mechanism for studying ICL over synthetic numerical sequences. Each transformer block uses multi-head self-attention. Following Garg et al.~\cite{garg2022template}, our medium model uses 8 attention heads, preserving the same head dimension-to-embedding dimension ratio. This ensures that differences in performance between architectures arise from depth / width changes rather than altered head geometry.

To assess whether the Garg architecture is appropriate for our setting (particularly given that our input dimensionality differs from theirs), we conducted a targeted architectural hyperparameter sweep. Garg et al.~\cite{garg2022template} work with inputs in a 10-50 dimensional vector space, whereas our examples are two-dimensional. This lets us check whether the same architecture scales appropriately to our lower-dimensional setting and whether different model capacities behave differently under our data distribution.

To simplify this evaluation and isolate the effect of model capacity, we restricted these preliminary experiments to the two-dimensional linear function class. Training only on linear functions avoids additional variability introduced by mixing function classes and provides a clean, controlled environment for assessing how architectural choices influence ICL dynamics. We also reduced the training to 100,000 steps (rather than the 500,000 steps used by Garg et al.~\cite{garg2022template}) because all models in our setting converged well before 100,000 steps, after which the loss curves exhibited small fluctuations but no further meaningful improvements. Table \ref{tab:model_sweep} summarizes the architectural configurations we evaluated.

\begin{table}[h]
\centering

\begin{tabular}{
    >{\centering\arraybackslash}m{1.5cm}
    >{\centering\arraybackslash}m{1.5cm}
    >{\centering\arraybackslash}m{3cm}
    >{\raggedright\arraybackslash}m{6cm}
}
\toprule
\textbf{Embeddings} & \textbf{Layers} & \textbf{Avg. Loss} & \textbf{Justification} \\
\midrule

128 & 8 & 0.054 &
Reduced-capacity model: half the embedding dimension and a shallower depth for a controlled low-capacity baseline.
\\
\addlinespace[10pt]

256 & 12 & 0.054 &
Matches Garg et al.~\cite{garg2022template}; serves as our reference architecture. \\
\addlinespace[10pt]

512 & 16 & 0.055 &
Higher-capacity model: doubles width and increases depth to test whether additional capacity improves ICL stability.
\\

\bottomrule
\end{tabular}

\caption{Architectures evaluated}
\label{tab:model_sweep}
\end{table}

After evaluating model capacity, we next specify the non-architectural hyperparameters used throughout the sweep. To maintain comparability with Garg et al.~\cite{garg2022template} while ensuring stable training for all model sizes, we adopt the hyperparameters listed in Table \ref{tab:training_hparams}.

\begin{table}[h]
\centering
\begin{tabular}{
    >{\centering\arraybackslash}m{3cm}
    >{\centering\arraybackslash}m{3cm}
    >{\raggedright\arraybackslash}m{6cm}
}
\toprule
\textbf{Hyperparameter} & \textbf{Value} & \textbf{Justification} \\
\midrule

Learning rate & $1\times10^{-4}$ &
Matches the setup of Garg et al.~\cite{garg2022template} and provided stable optimization across all model sizes. \\
\addlinespace[10pt]
Batch size & 64 &
Large enough for stable gradient estimates while remaining feasible across all architectures and the GPU used. \\
\addlinespace[10pt]
Optimizer & Adam &
Standard choice for transformer training; matches Garg et al.~\cite{garg2022template} \\
\addlinespace[10pt]
Points curriculum &
11 $\rightarrow$ 41 (add 2 every 2000 steps) &
Follows Garg et al.'s ~\cite{garg2022template} example-length schedule, enabling direct comparison while ensuring a gradually increasing in-context sequence length. \\

\bottomrule
\end{tabular}
\caption{Training hyperparameters used in our architectural sweep}
\label{tab:training_hparams}
\end{table}

To select an architecture for all subsequent experiments, we evaluated each model using the average loss over 100,000 training steps. Although the small and medium models achieved nearly identical average losses, the medium model displayed slightly smoother and faster convergence. Because our primary objective is to compare training curricula and study in-context adaptation, we chose the medium architecture as the best balance between robustness and capacity. We therefore adopt this model, together with the hyperparameters summarized in Table \ref{tab:training_hparams} for all remaining experiments.

%% file: Sections/Results.tex
\section{Results}

\subsection{Baselines}

We include two baselines. First, we evaluate a no-interruption setting, where the model receives only quadratic examples before the final quadratic query ($n_l = 0$, $n_q \geq 0$). This baseline shows the ideal quadratic performance in the absence of misleading context. It defines the target level of error that the recovery trajectories should move towards when the additional quadratic examples are provided after the linear priors. Second, we include the no-recovery setting, where the model receives linear priors but no quadratic examples after the switch ($n_l \geq 0$, $n_q = 0$). This baseline captures the model's error when it has no opportunity to recover from the misleading linear context. Together, these baselines characterize the best-case (no interruption) and worst-case (no recovery) regimes, allowing us to interpret the intermediate ($n_l$, $n_q$) configurations.

\subsection{Linear to Quadratic Context Switching}

Our inference tests sweep over two quantities: the number of linear examples shown before the task switch and the number of quadratic examples shown afterward. By varying these two values, we obtain a grid of ($n_l$, $n_q$) settings that reveals how strongly the preceding linear examples bias the model's predictions and how additional quadratic examples help the model recover. Figures 1-4 show how error changes across training curricula (random, sequential, and mixed) highlighting how context stickiness, recovery opportunities, and training strategy shape the model's behavior under context interruption.

\begin{figure}[h]
    \centering
    \includegraphics[width=0.8\linewidth]{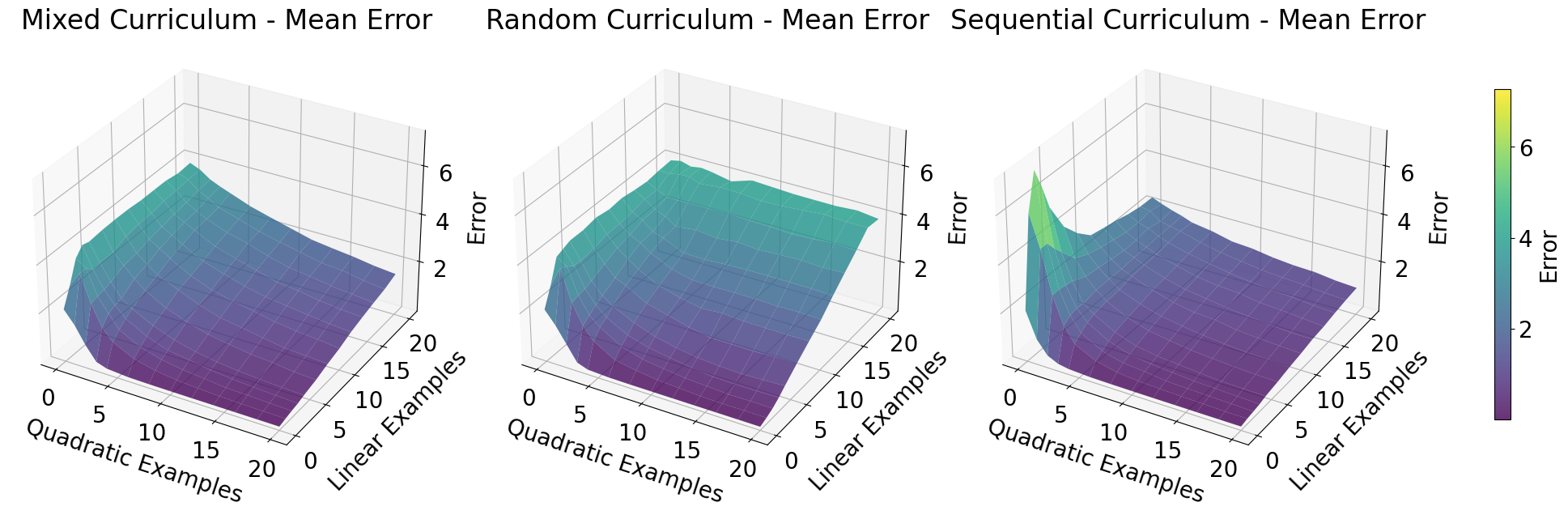}
    \caption{Overall 3D Error Surfaces}
    \label{fig:3d-error}
\end{figure}

Figure \ref{fig:3d-error} shows three-dimensional error surfaces for random, sequential, and mixed curricula, all plotted on the same scale. The horizontal axes indicate (1) how many linear examples appear before the task switch and (2) how many quadratic examples follow it. Across all curricula, two clear patterns emerge:

\begin{enumerate}
    \item Error grows as the number of preceding linear examples increases, reflecting the stronger influence of the earlier linear context.
    \item Error declines as more quadratic examples are provided, since additional post-switch evidence helps the model realign itself to the correct task.
\end{enumerate}

\begin{figure}[h]
    \centering
    \includegraphics[width=0.95\linewidth]{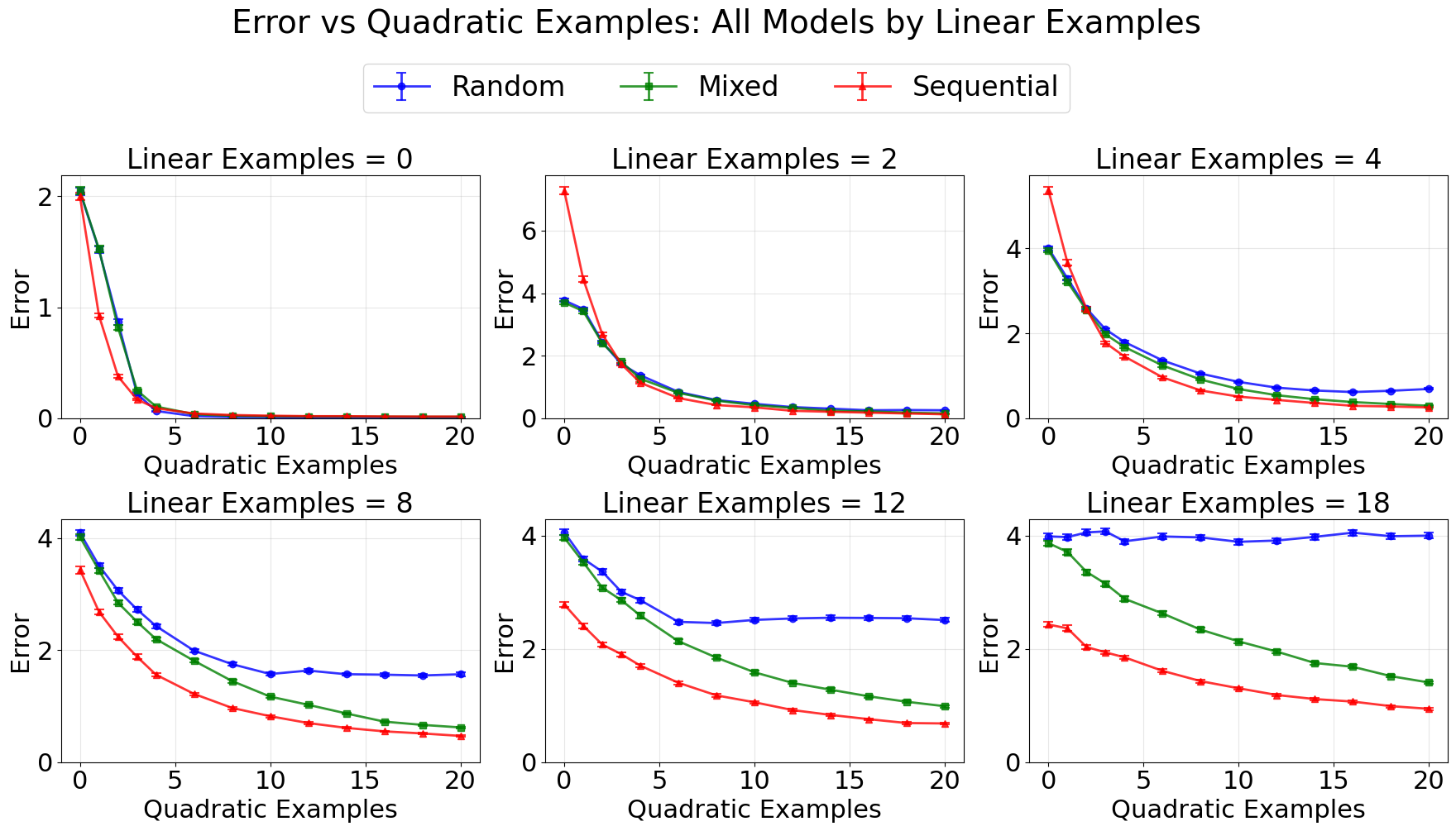}
    \caption{Recovery curves: holding the number of preceding linear examples constant}
    \label{fig:recovery_curves}
\end{figure}

Figure \ref{fig:recovery_curves} isolates the recovery behavior by fixing the number of preceding linear examples and varying the number of quadratic examples that follow. This view makes it easier to compare how quickly the model recovers across different curricula. The error bars show the standard error of the mean (SEM), which we use instead of standard deviation because our goal is to quantify uncertainty in the mean error across trials.

All curricula exhibit a recovery trend, but the rate of recovery differs substantially. Sequentially trained models recover the fastest, even when there are many preceding linear examples. Mixed curriculum models also improve with additional quadratic examples, though more gradually. In contrast, randomly trained models recover the least: once the number of linear examples becomes large, additional quadratic examples yield only minor improvements. This suggests that random training makes the model much more susceptible to misleading early context.

A diminishing-returns effect also appears across curricula. After a few quadratic examples, each extra example contributes less to reducing error. This diminishing effect becomes even weaker when the model has been exposed to a long sequence of linear examples, indicating that stronger initial bias requires more quadratic evidence to overcome.

\begin{figure}[h]
    \centering
    \includegraphics[width=0.95\linewidth]{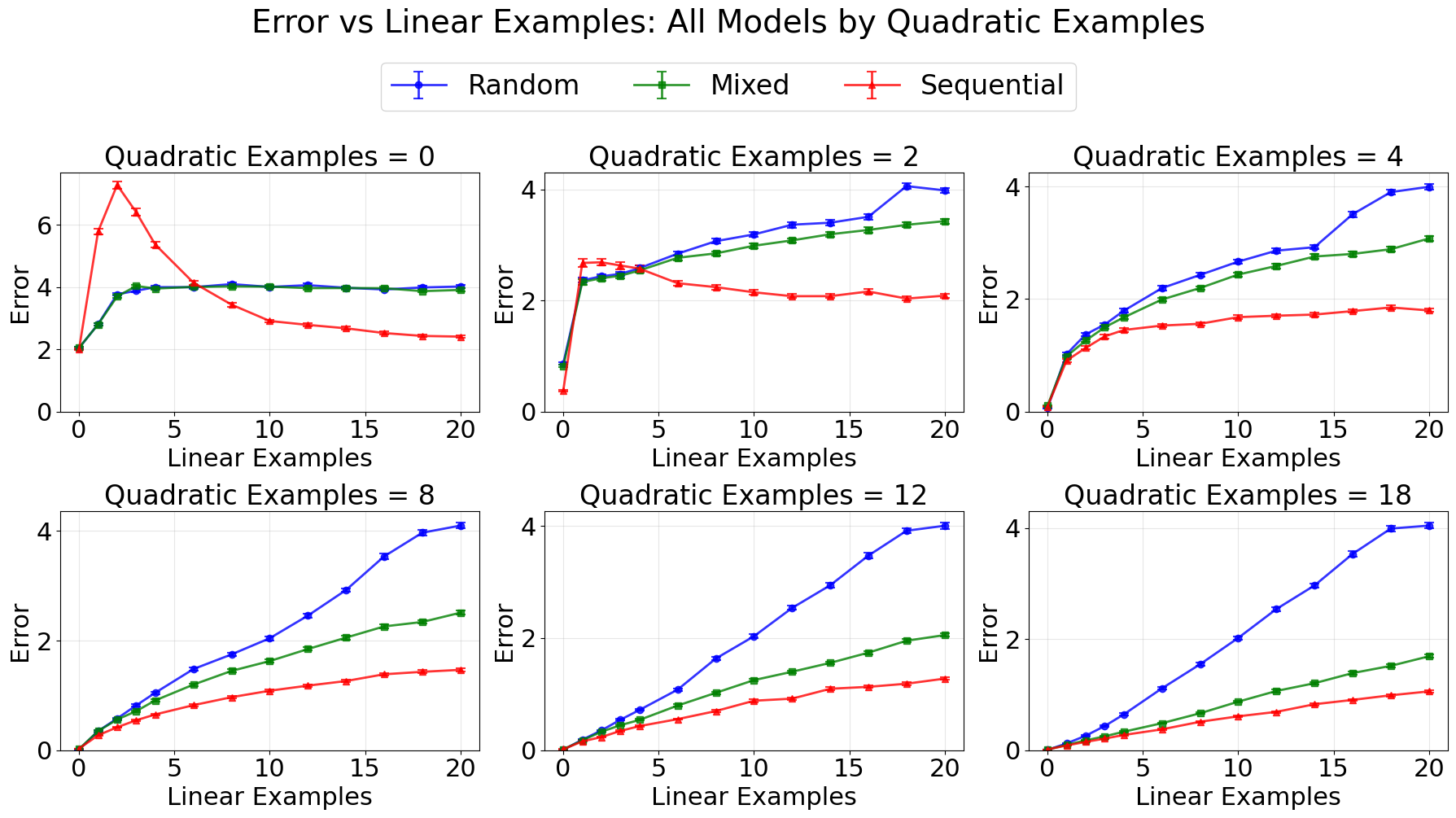}
    \caption{Stickiness curves: holding the number of following quadratic examples constant}
    \label{fig:stickiness_curves}
\end{figure}

Figure \ref{fig:stickiness_curves} fixes the number of quadratic examples and varies the number of linear examples shown before the switch. The resulting curves illustrate how error grows as the model is exposed to more linear context.

When the model receives enough quadratic examples after the switch (more than 2) and no preceding linear examples, the error is almost always zero, indicating that in a clean, uninterrupted setting the model can learn the quadratic mapping accurately. When there are fewer recovery quadratic examples, however, the first few linear examples cause disproportionately large increases in error, while additional linear examples contribute much smaller marginal effects. As the number of quadratic examples increases, the relationship between the number of linear examples and the resulting error becomes more linear. These patterns reveal how the model accumulates bias from earlier context and how effectively it can overcome this bias depending on the number of recovery examples available.

\begin{figure}[h]
    \centering
    \includegraphics[width=0.6\linewidth]{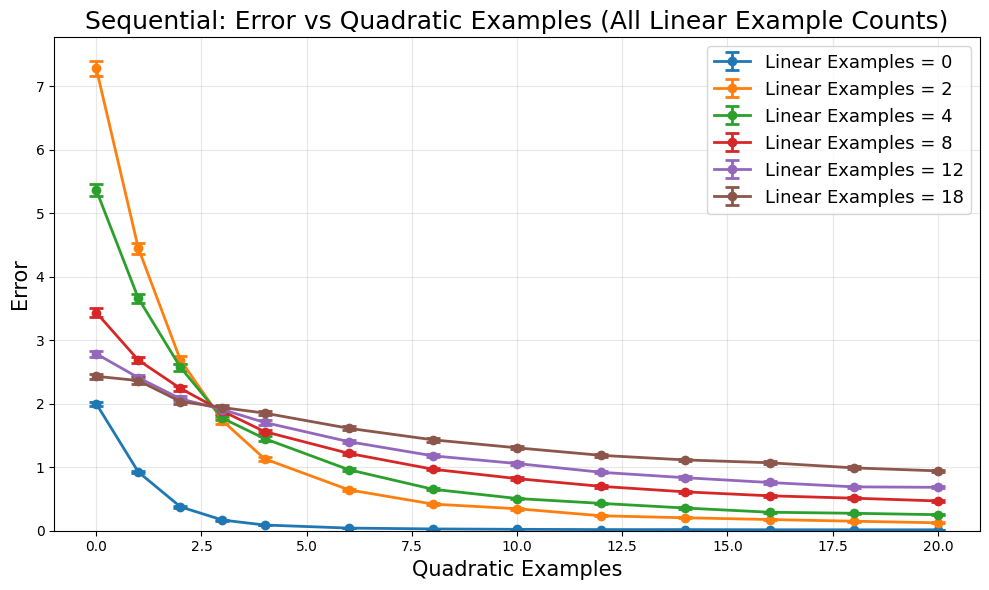}
    \caption{Sequential: error versus quadratic examples}
    \label{fig:seq_err_quad}
\end{figure}

Figure \ref{fig:seq_err_quad} examines the best-performing model (the sequentially trained model) and shows how its error changes as the number of quadratic examples increases, with each curve corresponding to a fixed number of preceding linear examples. This view allows us to compare recovery behavior under different amounts of prior linear context. A notable crossover effect emerges when only a very small number of quadratic examples are available. In this low-evidence regime, adding a small number of linear examples can actually reduce error. Intuitively, we believe this is a result of the linear points biasing the model towards quadratics with smaller second-order coefficients. Once enough quadratic examples are present, this effect disappears, and additional linear examples degrade performance.

\subsection{Comparing to Quadratic to Linear Context Switching}

\begin{figure}[h]
    \centering
    \includegraphics[width=0.95\linewidth]{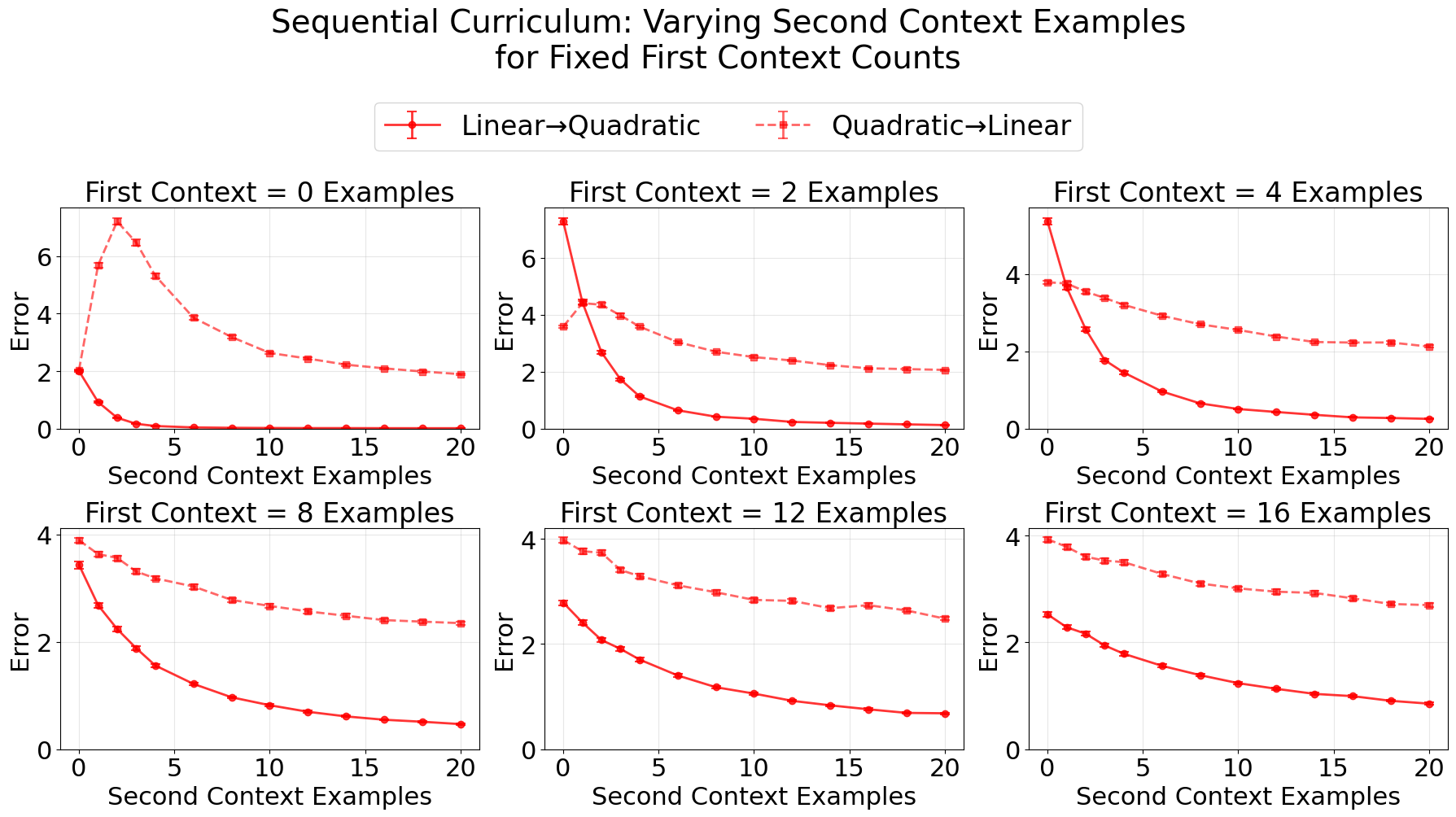}
    \caption{Sequential curriculum: error when switching between tasks}
    \label{fig:seq_quad_vs_lin}
\end{figure}

\begin{figure}[h]
    \centering
    \includegraphics[width=0.95\linewidth]{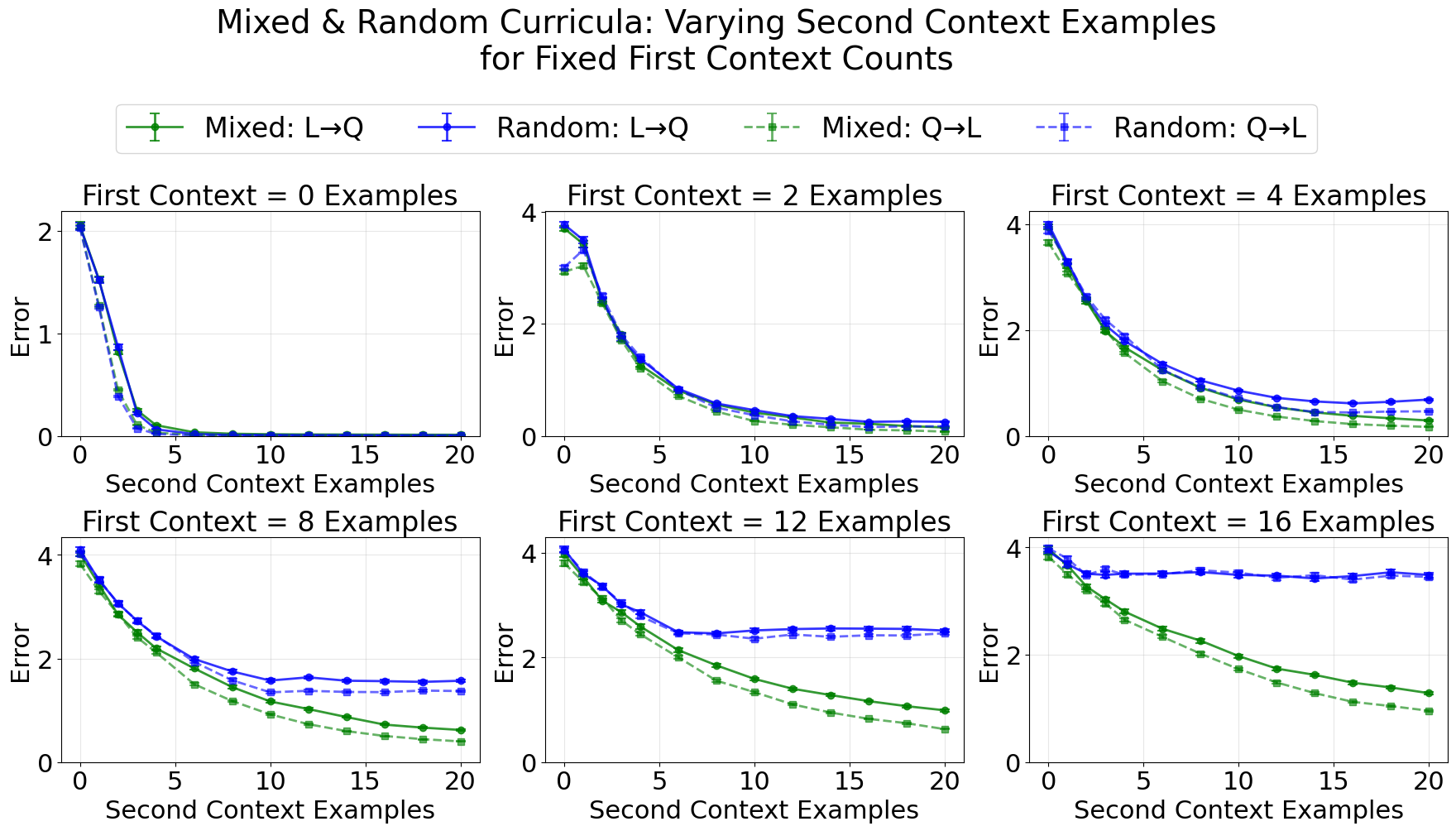}
    \caption{Mixed and random curricula: error when switching between tasks}
    \label{fig:mr_quad_vs_lin}
\end{figure}

So far, we have evaluated the setting in which our model must adjust to a linear to quadratic context switch. To check whether our observations depend on which task appears last in the sequence, we also evaluate the reverse setting, where the model must predict a linear example after being given quadratic examples. Figures \ref{fig:seq_quad_vs_lin} and \ref{fig:mr_quad_vs_lin} show the error as a function of the second context length for fixed counts of the first context, and include curves for both linear to quadratic and quadratic to linear switches.

Across curricula, we see the same high-level pattern in both directions: by keeping the number of second context examples fixed, more examples from the first task lower performance, while increasing the number of second task examples helps the model recover, confirming that these effects hold true regardless of the task we evaluate on.

However, the curricula differ sharply in how asymmetric the two directions become. For the sequential model, switching from linear to quadratic consistently yields much lower error than switching from quadratic to linear. Since this model is trained on linear functions in the first half and exclusively on quadratics in the second half, this gap suggests a form of catastrophic forgetting: it remains strong on the task it trained on last (quadratic), but becomes noticeably worse at predicting linear functions once training has shifted away from them.

In contrast, the mixed and random models show similar performance for linear to quadratic and quadratic to linear sequences with mixed exhibiting stronger recovery with longer initial context sequences. Although both tend to perform slightly better on the quadratic to linear setting, the error curves mostly follow each other. This indicates that these curricula do not suffer from the same level of catastrophic forgetting as sequential. It also reinforces our earlier conclusions: the stickiness and recovery patterns we observed in the linear to quadratic experiments are not an artifact of always evaluating on quadratic outputs. They persist when we flip the evaluation to linear predictions as well.

%% file: Sections/Conclusion.tex
\section{Conclusion}

We studied how transformers handle abrupt task switches in ICL by alternating linear and quadratic examples under different training curricula. Across sequential, mixed, and random training, we observed context stickiness: longer runs of misleading linear examples increased error on subsequent quadratic predictions, while additional quadratic examples improved recovery with diminishing returns. Sequentially trained models yielded the fastest and most complete recovery, random training was the most brittle, and mixed training fell between these two extremes. These interference and recovery patterns held in both linear-to-quadratic and quadratic-to-linear switches, although sequential training also produced catastrophic forgetting on the task seen earlier in training.

Our study is conducted in a controlled synthetic setting using small self-trained transformers. The results should therefore be interpreted as evidence about interference under task switching in simplified ICL settings, rather than as a direct quantitative claim about frontier pretrained LLM behavior.

Future work could compare these behaviors to closed-form regression baselines and explore architectural or prompting strategies that explicitly mitigate in-context interference.